\documentclass[acmtog]{acmart}
\setcopyright{none}
\usepackage{graphicx}
\usepackage{amsmath}
\usepackage{hyperref}
\usepackage[inline]{enumitem}
\usepackage{booktabs}
\usepackage[table]{xcolor}
\usepackage{microtype}
\providecommand{\correspondingauthor}{}

\AtBeginDocument{%
  }

\date{}

\begin{document}

\title{Rethinking Sparse-View 3D Gaussian Splatting:\\
A Few-Shot Learning Perspective on Overfitting Prevention}

\author{Lejia Ye}
\authornote{Equal contribution.}
\email{yelj8@mail2.sysu.edu.cn}
\affiliation{%
  \institution{School of Aeronautics and Astronautics, Sun Yat-sen University}
  \city{Shenzhen}
  \country{China}}
\author{Bin Wu}
\authornotemark[1]
\email{wubin58@mail2.sysu.edu.cn}
\affiliation{%
  \institution{School of Aeronautics and Astronautics, Sun Yat-sen University}
  \city{Shenzhen}
  \country{China}}
\author{Jingjing Gao}
\email{gaojj27@mail2.sysu.edu.cn}
\affiliation{%
  \institution{School of Aeronautics and Astronautics, Sun Yat-sen University}
  \city{Shenzhen}
  \country{China}}
\author{Yiyang Wu}
\email{wuyy239@mail2.sysu.edu.cn}
\affiliation{%
  \institution{School of Aeronautics and Astronautics, Sun Yat-sen University}
  \city{Shenzhen}
  \country{China}}
\author{Xiaohu Zhang}
\correspondingauthor
\email{zhangxiaohu@mail.sysu.edu.cn}
\affiliation{%
  \institution{School of Aeronautics and Astronautics, Sun Yat-sen University}
  \city{Shenzhen}
  \country{China}}
\renewcommand{\shortauthors}{Ye et al.}


\begin{abstract}
Sparse-view novel view synthesis (NVS) is essential for AR/VR and real-time communication applications, yet remains challenging when reconstructing 3D scenes from limited viewpoints. While 3D Gaussian Splatting (3DGS) excels in dense-view scenarios, it suffers from overfitting and geometry degradation under sparse inputs. 
Current approaches often tackle these problems from a local perspective, addressing either data quality or training regularization individually, lacking a unified and interpretable framework for systematic improvement.
We introduce a new perspective that reframes sparse-view 3DGS from a few-shot learning viewpoint, treating each Gaussian primitive as a neuron and leveraging deep learning regularization techniques. Under this unified framework that systematically connects 3DGS optimization with deep learning regularization principles, our approach generates more effective supervision from scarce observations through two strategies: (1) data-related supervision enrichment via optical flow-guided pseudo-view synthesis and joint camera–scene optimization; and (2) training-time regularization via motion-aware dropout and early stopping, which suppress overfitting while preserving dynamic-scene stability. Extensive experiments on the SIGGRAPH Asia 2025 volumetric video challenge sparse-view dataset demonstrate significant improvements over state-of-the-art optimization-based and feed-forward baselines.
Our system won first place in the sparse-view track of the SIGGRAPH Asia 2025 challenge.
\end{abstract}
\keywords{3D Gaussian Splatting, sparse-view synthesis, few-shot learning}
\maketitle

\ccsdesc[500]{Computing methodologies~Computer graphics}
\ccsdesc[500]{Computing methodologies~Rendering}
\ccsdesc[300]{Computing methodologies~Machine learning}
\ccsdesc[300]{Computing methodologies~Computer vision}


\begin{figure*}[t]
 \centering
 \includegraphics[width=0.96\textwidth]{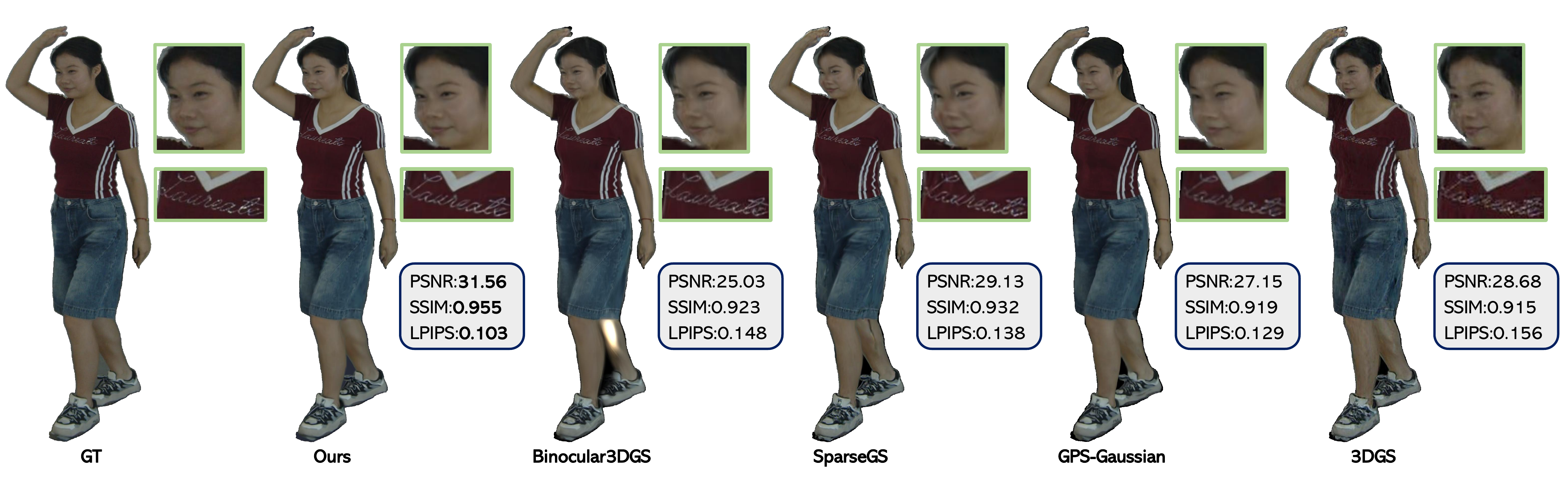}
 \caption{We reinterpret sparse-view 3D Gaussian Splatting through the lens of deep-learning overfitting prevention, design corresponding strategies, and outperform both state-of-the-art optimization-based (Binocular3DGS, SparseGS, 3DGS) and feed-forward (GPS-Gaussian) baselines.}
 \label{fig:teaser}
 \vspace{-0.2cm}
\end{figure*}


\vspace{-0.3cm}
\section{Introduction}

\begin{figure*}[htbp]
	\centering
\includegraphics[width=0.98\textwidth]{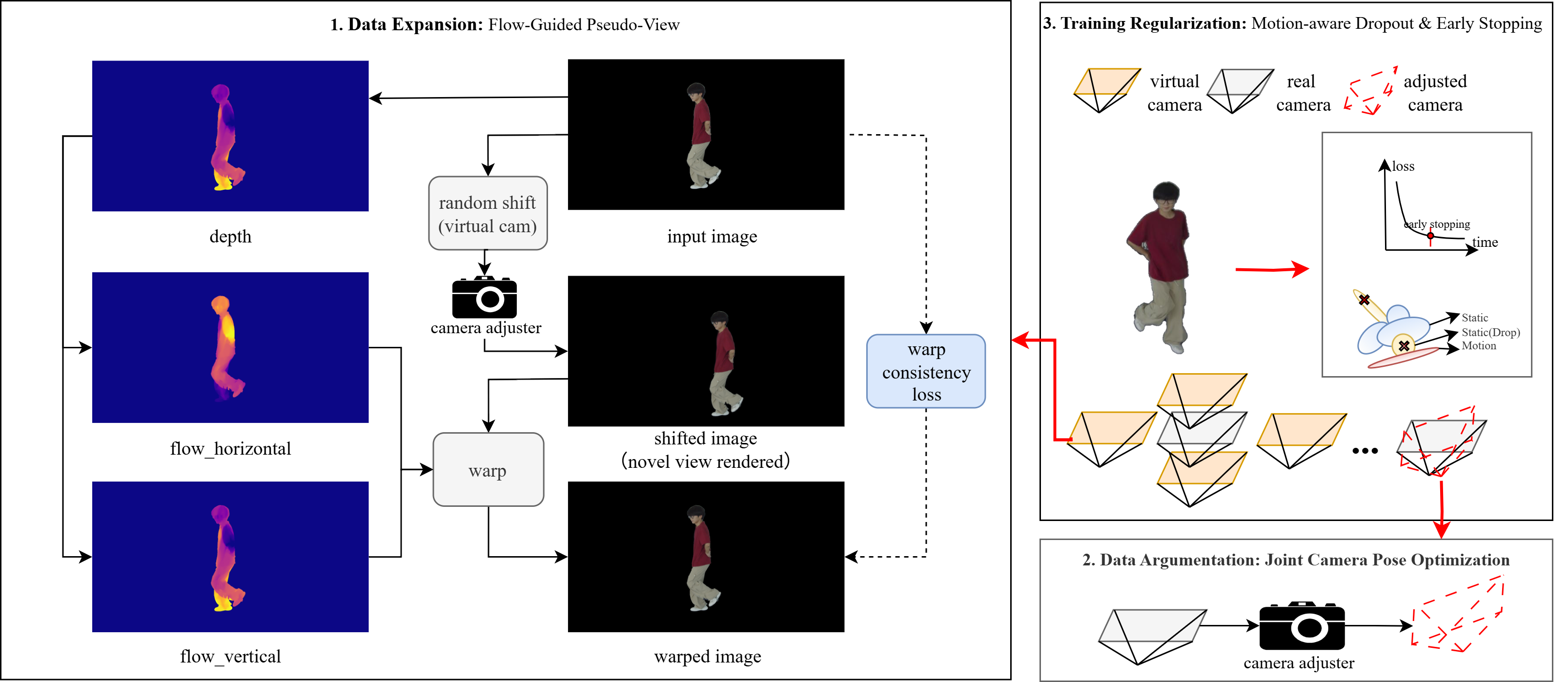}
	\caption{Overview of our methodology. Data Expansion: Depth plus horizontal/vertical optical flow produce virtual camera shifts whose rendered views are warped back and constrained by a consistency loss; Data Augmentation: Camera poses with large reprojection error are corrected through a joint camera–scene optimizer to suppress calibration noise; Training Regularization: Motion-aware dropout acts only on static Gaussians and is coupled with validation-selected early stopping to regularize training.}
	\label{fig:pipeline}
	\vspace{-0.5cm}
\end{figure*}

Sparse-view novel view synthesis (NVS) aims to reconstruct a renderable 3D scene from a small number of input viewpoints, a capability that is critical for AR/VR and real-time communication applications. 
Recent advances in differentiable rendering—most notably Neural Radiance Fields (NeRF) and 3D Gaussian Splatting (3DGS)—have driven substantial progress; in dense-view settings, 3DGS in particular offers fast convergence and high rendering efficiency and has been widely adopted \cite{kerbl20233d}. However, under sparse inputs, even 3DGS often overfits the limited viewpoints, resulting in geometry degradation and unstable, artifact-prone appearance.

Existing approaches address the sparse viewpoint problem through various strategies—synthesizing pseudo-views, incorporating external depth supervision, and applying training regularization \cite{bao2024loopsparsegs,yu2024lm,li2024dropgaussian,smith2024coadaptation}—but lack a unified perspective. 
This motivates us to seek a principled viewpoint that organizes these strategies within a single methodology. We therefore ask: \textbf{\emph{if each Gaussian primitive could be treated as a neuron, can we borrow the well-established regularization toolkit from deep learning?}} This question leads us to reframe sparse-view 3DGS from a few-shot learning perspective, making deep learning regularization principles directly applicable.

In deep learning, Ying et al.\ \cite{ying2019overview} attribute overfitting to three causes—noisy data, excessive model capacity, and insufficient data. We argue that a corresponding mapping to sparse-view 3DGS challenges is: camera miscalibration introduces noise, excessive and entangled Gaussians lead to overfitting and co-adaptation, and limited viewpoints provide insufficient supervision. Correspondingly, the four canonical overfitting prevention strategies in deep learning—early stopping, model reduction, data augmentation, and regularization—naturally apply to 3DGS to address these challenges. We therefore group them into the two pillars that structure our method: \emph{Supervision Enhancement} (data-related) and \emph{Training Regularization} (model-training-related).

Our approach instantiates these pillars through: (1) \emph{Supervision Enhancement}: data expansion via optical flow-guided pseudo-view synthesis and data augmentation via joint camera–scene optimization; (2) \emph{Training Regularization}: motion-aware dropout (restricted to static regions to handle dynamic scenes) paired with early stopping. 

\noindent\textbf{Contributions.} Our work makes the following contributions:
\begin{itemize}[leftmargin=*]
\vspace{-0.15cm}
\item To the best of our knowledge, we are the first to provide a principled reframing of sparse-view 3D Gaussian Splatting as a few-shot learning problem by interpreting each Gaussian primitive as a computational unit. This perspective yields an interpretable link between graphics optimization and well-established generalization mechanisms in deep learning.
\item We instantiate the \emph{Supervision Enhancement} component via a supervision scheme tailored to limited viewpoints that (i) synthesizes pseudo-views through optical flow, and (ii) performs adaptive joint camera–scene optimization based on reprojection error.
\item We instantiate the \emph{Training Regularization} component by coupling motion-aware dropout—selectively applied to static Gaussians—with early stopping, effectively reducing co-adaptation under sparse inputs.
\item Experiments on the SIGGRAPH Asia 2025 sparse-view dataset demonstrate improvements over both feed-forward and optimization-based methods across quantitative metrics and visual quality.
\vspace{-0.4cm}
\end{itemize}

\section{Methodology}

Our overall algorithm is illustrated in Figure \ref{fig:pipeline}. 
Following the deep learning perspective outlined in the introduction, we organize our approach into two complementary categories: \emph{Supervision Enhancement} and \emph{Training Regularization}. This unified framework addresses the three fundamental causes of overfitting in sparse-view 3DGS: noisy data (corresponding to camera miscalibration), insufficient data, and excessive model capacity. In the following subsections, we detail each component.
\vspace{-0.3cm}
\subsection{Supervision Enhancement}
To address insufficient supervision and noisy data in sparse-view 3DGS, we enhance the training data through two strategies: pseudo-view synthesis addresses insufficient data, while joint camera–scene optimization addresses noisy data.
\vspace{-0.3cm}
\subsubsection{Optical Flow Guided Pseudo-view Synthesis}\label{sec:pseudo-view}
Given a predicted depth map $D$ with intrinsics $(f_x,f_y,c_x,c_y)$, we synthesize a nearby view by applying a small in-plane translation $\boldsymbol{t}=(t_x,t_y,0)^\top$. The induced pixel shifts are
\begin{align}
u(\boldsymbol{x})=\frac{f_x(-t_x)}{D(\boldsymbol{x})},\qquad
v(\boldsymbol{x})=\frac{f_y(-t_y)}{D(\boldsymbol{x})}.
\end{align}
Rendering the translated pose produces $\widetilde{\boldsymbol{I}}$, which we warp back via differentiable sampling:
\begin{align}
\boldsymbol{I}_w(\boldsymbol{x})=\widetilde{\boldsymbol{I}}\!\left(\boldsymbol{x}+\begin{bmatrix}u(\boldsymbol{x})\\ v(\boldsymbol{x})\end{bmatrix}\right).
\end{align}
This optical-flow parameterization explicitly reasons about horizontal and vertical displacements, providing extra robustness along the vertical axis when the training views lie roughly on the same horizontal ring (a common sparse-view setting where vertical supervision is weak). Over the valid pixel set $\mathcal{V}$ (visible and in-bounds), we measure the Charbonnier photometric consistency as
\begin{align}
\mathcal{L}_{\text{photo}}
=\frac{1}{|\mathcal{V}|}\sum_{\mathbf{x}\in\mathcal{V}}
\sqrt{\left\|\mathbf{I}_w(\mathbf{x})-\mathbf{I}_{\text{gt}}(\mathbf{x})\right\|_2^2+\epsilon},
\end{align}
where $\epsilon>0$ is a small constant.

To regularize depth and the induced flow, we adopt total variation (TV) regularization for both depth and flow fields, along with edge-aware smoothing to preserve depth discontinuities at image edges. The final geometric objective combines the photometric consistency loss with these regularization terms:
\begin{align}
\mathcal{L}_{\text{geo}} ={} & \lambda_{\text{geo}}\mathcal{L}_{\text{photo}} \\
&+\lambda_{d}\mathcal{L}_{\text{depth-TV}}
+\lambda_{f}\mathcal{L}_{\text{flow-TV}}
+\lambda_{e}\mathcal{L}_{\text{edge}},
\end{align}
where $\mathcal{L}_{\text{depth-TV}}$ and $\mathcal{L}_{\text{flow-TV}}$ are total variation terms that encourage smoothness in depth and flow fields, respectively, and $\mathcal{L}_{\text{edge}}$ is an edge-aware term that reduces smoothing at image edges. The hyperparameters are $\lambda_{\text{geo}},\lambda_{d},\lambda_{f},\lambda_{e}>0$.

\subsubsection{Joint Camera–Scene Optimization}\label{sec:camera-opt}
Camera poses are part of the input data; when their calibration is inaccurate, they act as noise that degrades sparse-view reconstructions. To address this, we jointly optimize both camera poses and scene geometry based on reprojection error. We parameterize camera poses using $\mathrm{SO}(3)$ rotations and $\mathbb{R}^3$ translations (SO3xR3), allowing the optimization to correct calibration errors while simultaneously refining the 3D Gaussian scene representation. To avoid unnecessary optimization when camera calibration is already accurate, we selectively enable camera pose optimization only for sequences with high reprojection errors, which indicates potential calibration issues. The camera optimizer is activated after an initial warmup period and frozen before geometric constraints are applied, ensuring stable convergence. This selective joint optimization effectively reduces the impact of noisy camera parameters on sparse-view reconstruction while preserving well-calibrated camera poses.

\vspace{-0.2cm}
\subsection{Training Regularization}
To address excessive model capacity and prevent overfitting, we employ two regularization strategies: motion-aware dropout and early stopping.
\vspace{-0.2cm}
\subsubsection{Motion-Aware Dropout}\label{sec:dropout}
Dropout requires that different subsets of Gaussians, after being dropped, yield comparable outcomes. However, this consistency assumption is often violated under fast motion. We therefore restrict dropout to static regions only. To define static and dynamic regions, we leverage dense initialization that combines optical flow estimation (PDCNet\cite{truong2023pdc}) and multi-view point cloud reconstruction (OpenMVS\cite{openmvs_github}). PDCNet provides a complete point cloud covering both static and dynamic regions, while OpenMVS, relying on multi-view consistency, primarily produces static point clouds. We compute the distance from each point in the PDCNet point cloud to the nearest point in the OpenMVS point cloud. Points close to OpenMVS (within a threshold) are assigned as static, while distant points are labeled as dynamic.

During training, we apply motion-aware dropout: for each Gaussian, we determine whether it lies in a dynamic region based on the initial definition. Gaussians in dynamic regions are excluded from dropout to preserve rendering quality, while only static-region Gaussians are subject to dropout regularization. This selective application maintains the consistency assumption for dropout while protecting dynamic regions.
\vspace{-0.2cm}
\subsubsection{Early Stopping}\label{sec:early-stopping}
To prevent overfitting, we employ early stopping based on validation metrics. We determine the optimal early-stopping point by evaluating checkpoints throughout training and computing rankings on novel view synthesis metrics (PSNR, SSIM, LPIPS). The optimal stopping point is determined as the checkpoint with the lowest average rank across all three metrics. This stopping criterion is determined using a validation sequence and then applied consistently across all experiments.

\begin{table}[t]
	\vspace{-0.5cm}
	\centering
	\caption{Quantitative comparison using PSNR, SSIM, and LPIPS on the validation set (metrics computed on the first 50 frames of each sequence). Best results are highlighted with red background, and second-best with orange background. Methods marked with * are trained at half resolution and upsampled to original resolution for evaluation.}
	\label{tab:quant_metrics}
\begingroup
\setlength{\tabcolsep}{3pt}
\renewcommand{\arraystretch}{1.08}
\scriptsize
\resizebox{\columnwidth}{!}{%
\begin{tabular}{@{}l l rrr rrr@{}}
		\toprule
		Method & Pub & \multicolumn{3}{c}{Seq. 012} & \multicolumn{3}{c}{Seq. 001} \\
		 & & PSNR$\uparrow$ & SSIM$\uparrow$ & LPIPS$\downarrow$ & PSNR$\uparrow$ & SSIM$\uparrow$ & LPIPS$\downarrow$ \\
		\midrule
		3DGS* \cite{kerbl20233d} & TOG'23 & 24.36 & 0.893 & 0.167 & \cellcolor{orange!30}{25.87} & 0.898 & 0.193 \\
		Binocular3DGS \cite{han2024binocular} & NeurIPS'24 & \cellcolor{orange!30}{24.68} & \cellcolor{orange!30}{0.908} & 0.151 & 25.77 & \cellcolor{orange!30}{0.909} & 0.175 \\
		GPS-Gaussian \cite{zheng2024gps} & CVPR'24 & 24.40 & 0.896 & \cellcolor{orange!30}{0.149} & 24.54 & 0.890 & \cellcolor{orange!30}{0.170}  \\
		SparseGS* \cite{SparsegsXiong2025} & 3DV'25 & 24.27 & 0.901 & 0.155 & 25.24 & 0.896 & 0.189  \\
		Ours & -- & \cellcolor{red!30}{26.55} & \cellcolor{red!30}{0.923} & \cellcolor{red!30}{0.137} & \cellcolor{red!30}{27.13} & \cellcolor{red!30}{0.924} & \cellcolor{red!30}{0.143} \\
		\bottomrule
	\end{tabular}
}%
\endgroup
\vspace{-0.2cm}
\end{table}

\vspace{-0.2cm}
\subsection{Training Objective and Optimization}

The overall training objective combines rendering loss, geometric objective, and camera pose regularization. We optimize training under sparse views by combining early stopping and motion-aware dropout to stabilize convergence and improve efficiency.

\noindent\textbf{Rendering Loss}
Let $\hat{\mathbf{I}}$ be the rendered image and $\mathbf{I}_{\text{gt}}$ the ground truth. We adopt a standard combination of $\ell_1$ and SSIM:
\begin{align}
\mathcal{L}_{\text{render}}
= \lambda_{1}\,\|\hat{\mathbf{I}}-\mathbf{I}_{\text{gt}}\|_{1}
+ \lambda_{\text{SSIM}}\,\big(1-\text{SSIM}(\hat{\mathbf{I}},\mathbf{I}_{\text{gt}})\big).
\end{align}

\noindent\textbf{Geometric Objective}
The geometric objective $\mathcal{L}_{\text{geo}}$ combines photometric consistency with depth and flow regularization, as detailed in Section~\ref{sec:pseudo-view}.

\noindent\textbf{Camera Regularization}
The camera regularization term $\mathcal{L}_{\text{cam}}$ constrains the deviation of optimized camera poses from their initial values, preventing excessive pose modifications.

\noindent\textbf{Overall Objective}
The complete training objective is a weighted sum of all three components:
\begin{align}
\mathcal{L}_{\text{total}} = \lambda_{\text{render}}\mathcal{L}_{\text{render}} + \lambda_{\text{geo}}\mathcal{L}_{\text{geo}} + \lambda_{\text{cam}}\mathcal{L}_{\text{cam}},
\end{align}
where $\lambda_{\text{render}}$, $\lambda_{\text{geo}}$, and $\lambda_{\text{cam}}$ are the weights for the rendering loss, geometric objective, and camera regularization, respectively. 

\vspace{-0.2cm}
\begin{figure*}[!t]
\centering
\includegraphics[width=0.98\textwidth]{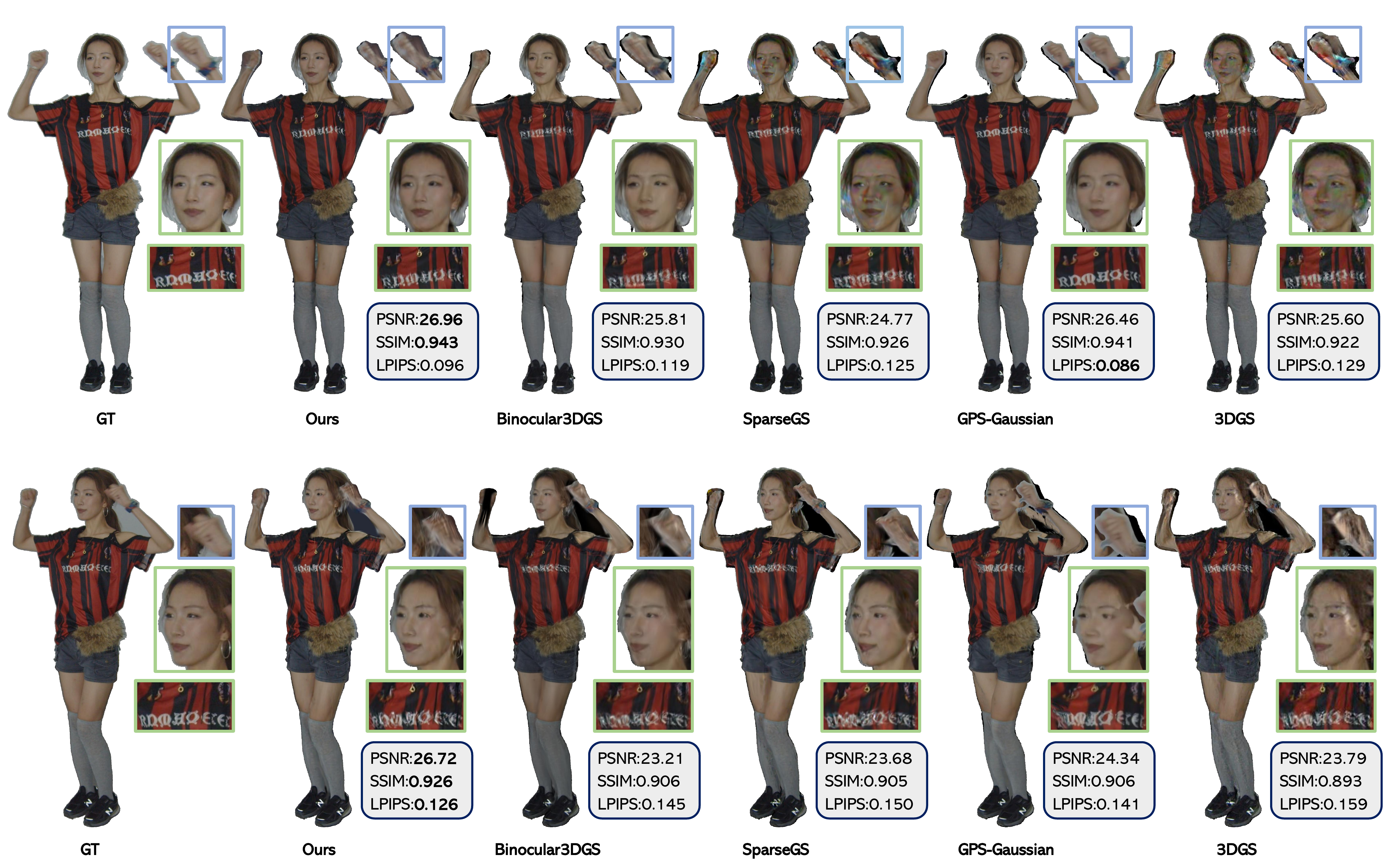}
\caption{Qualitative comparison on challenging sequences with fast human motion. Our method produces more stable, artifact-free reconstructions, especially in dynamic regions. Entries marked with * use half-resolution training and are resized for evaluation.}
\label{fig:012comparefig}
\end{figure*}

\section{Experiments}

\subsection{Experimental Settings}

\noindent\textbf{Dataset and Evaluation Protocol}
We evaluate our method on the SIGGRAPH Asia 2025 Volumetric Video Challenge sparse-view track dataset \cite{sigasia2025volumetric}. The dataset consists of 2 validation sequences (with full 16 camera viewpoints) and 5 test sequences (with 8 camera viewpoints), each containing 300 frames. All quantitative comparisons reported in this paper are conducted on the validation sequences, where we use 8 training views for reconstruction and evaluate on 8 novel test views. Test sequences are reserved for challenge submissions. Metrics (PSNR, SSIM, LPIPS) are computed on foreground regions only across all novel views.

\noindent\textbf{Implementation Details}
All experiments were conducted on an NVIDIA A100 40GB GPU, with the training phase of each run taking roughly three minutes (preprocessing time is excluded). The training process consisted of 5,000 epochs with a progressive resolution schedule: we trained 3,000 epochs at half resolution followed by 2,000 epochs at full resolution. Although the nominal schedule budgets 30k optimization steps, we monitor PSNR/SSIM/LPIPS on a held-out validation sequence and consistently trigger early stopping around step~5k.

\noindent\textbf{Hyperparameters}
For the rendering loss, we set $\lambda_1=0.8$ and $\lambda_{\text{SSIM}}=0.2$. The geometric objective weights are: $\lambda_{\text{geo}}=0.8$, $\lambda_d=0.05$ (depth TV), $\lambda_f=0.1$ (flow TV), and $\lambda_e=0.05$ (edge-aware). Camera optimization is enabled for sequences with reprojection error exceeding 1.0, activated at step 3,000 and frozen at step 4,000. Motion-aware dropout uses a factor of 0.1 for static Gaussians, with motion detection threshold of 0.8. The geometric constraint starts at iteration 500 with translation range $(0.05, 0.2)$.

\vspace{-0.3cm}
\subsection{Comparison with SOTA Methods}

Our comparative study covers state-of-the-art optimization-based and feed-forward methods, specifically:
\begin{enumerate*}[label=\arabic*]
\item the base 3DGS method \cite{kerbl20233d};
\item a representative self-supervised pseudo-data approach, Binocular3DGS \cite{han2024binocular};
\item a representative feed-forward method, GPS-Gaussian \cite{zheng2024gps};
\item the state-of-the-art for sparse reconstruction, SparseGS \cite{SparsegsXiong2025}.
\end{enumerate*}
Quantitative comparisons are reported in Table~\ref{tab:quant_metrics}, while qualitative results are visualized in Figures~\ref{fig:teaser} and \ref{fig:012comparefig}.

\vspace{-0.4cm}
\section{Conclusion}
\noindent\textbf{Conclusion}
We reframe sparse-view 3D Gaussian Splatting from a few-shot learning perspective and organize the solution around two pillars: \emph{Supervision Enhancement} (optical-flow pseudo views plus joint camera–scene optimization) and \emph{Training Regularization} (motion-aware dropout plus early stopping). Together with a stabilized training objective, this unified methodology delivers consistent quantitative and qualitative gains on the SIGGRAPH Asia 2025 sparse-view benchmark.

\noindent\textbf{Limitations and Future Work}
The current pipeline still assumes moderate motion when synthesizing pseudo views and extremely rapid human dynamics can violate this assumption. Future work will explore explicit motion compensation or learned pose priors to better align highly dynamic sequences.

\bibliographystyle{ACM-Reference-Format}
\bibliography{references}

\end{document}